# Towards an Ontology based integrated Framework for Semantic Web

Nora Y. Ibrahim

Computer and System Department,
Electronic Research Institute
Cairo, Egypt
nora@eri.sci.eg

Sahar A. Mokhtar

Computer and System Department,
Electronic Research Institute
Cairo, Egypt
sahar@eri.sci.eg

Hany M. Harb

Computer and Systems Engineering
Department, Faculty of Engineering,
Al-Azhar University
Cairo, Egypt
harbhany@yahoo.com

*Abstract*—**This Ontologies are widely used as a means for solving the information heterogeneity problems on the web because of their capability to provide explicit meaning to the information. They become an efficient tool for knowledge representation in a structured manner. There is always more than one ontology for the same domain. Furthermore, there is no standard method for building ontologies, and there are many ontology building tools using different ontology languages. Because of these reasons, interoperability between the ontologies is very low. Current ontology tools mostly use functions to build, edit and inference the ontology. Methods for merging heterogeneous domain ontologies are not included in most tools. This paper presents ontology merging methodology for building a single global ontology from heterogeneous eXtensible Markup Language (XML) data sources to capture and maintain all the knowledge which XML data sources can contain.**

## I. INTRODUCTION

Ontologies have been realized as the key technology for shaping and exploiting information for the effective management of knowledge. The study of ontologies and their use is no longer just one of the fields in the Artificial Intelligence. Ontologies are now ubiquitous in many information-systems enterprises: they constitute the backbone for the Semantic web, they are used in E-commerce, and in various application fields such as E-science, digital libraries, bioinformatics and medicine. As a result, developers are designing a large number of ontologies using different tools and different languages. These ontologies cover unrelated or overlapping domains, at different levels of detail and granularity. Multiple ontologies need to be accessed from several applications. Such wide-spread use of ontologies inevitably produces an ontology-management problem. Ontology management is the whole set of methods, methodologies, and techniques that is necessary to efficiently use multiple variants of ontologies from possibly different sources for different tasks. Ontology management includes operations such as mapping, alignment, matching, integration and merging. Ontology mapping aims to find semantic correspondences between similar elements of different ontologies [1]. Ontology matching is the process of detecting links between entities in heterogeneous ontologies. Ontology alignment is the task of creating links between two original ontologies. Ontology alignment is made if the sources become consistent with each other but are kept separate [2]. Ontology alignment is made when they usually have complementary domains. Ontology integration is the process of building an ontology in one subject reusing one or more ontologies in different subjects [3]. Ontology merging is the process of generating a single, coherent ontology from two or more existing and different ontologies related to the same subject [3]. A merged single coherent ontology includes information from all source ontologies but is more or less unchanged. The original ontologies have similar or overlapping domains but they are unique and not revisions of the same ontology [4]. The merging process can be performed in a number of ways, manually, semi automatically, or automatically. Manual ontology merging is a difficult, time-consuming and error prone task due to the continuous growth in both the size and number of ontologies. Therefore, several automatic and semi-automatic ontology merging frameworks have recently been proposed. These frameworks aim to find semantic correspondences between the concepts of the ontologies for a specific domain through exploiting syntactic- and/or semantic-based techniques. A new method is presented for generation of OWL ontologies (local ontologies) from heterogeneous XML data sources [5]. This paper illustrates the process of merging these local ontolgies to create a global ontology which is the union of the source ontologies. The merged ontology captures all the knowledge from the original ontologies. The challenge in ontology merging is to ensure that all correspondences and differences between the ontologies are reflected in the merged ontology. Merging of local ontologies is performed using protégé. Protégé is a free, open-source ontology editor which supports two ways of modeling ontologies, namely Protégé-Frames and Protégé OWL [6] where Prompt plug-in/tab is used for merging. Section 2 discusses most common methodologies and tools used for ontology management. Section 3 illustrates the framework to generate a global ontology from heterogeneous XML data sources. Section 4 describes methodology for ontology merging process. This merging methodology is considered to be integration to the system presented in [5]. Section 4 focuses on experiment and results of merging process. Section 5 contains the conclusion and future work.

## II. RELATED WORK

This section introduces the current research in ontology management field. Section 2.1 discusses most common methodologies of ontology management while section 2.2 presents ontology management tools used in ontology merging

### A. Ontology Management Methodologies

Recently, there has been considerable interest in ontology management methodologies and techniques to assist in a variety of ontology management operations, e.g., mapping, merging and alignment. The following methodologies represent the current research in ontology merging field.

#### 1) Miyoung Cho and Pankoo Kim [7]

Miyoung Cho, Hanil Kim and Pankoo Kim proposed ontology merging method using vertical and horizontal approaches based in WordNet. They presented the problem of proximity between two ontologies as a choice between alignment and merging. The alignment is limited to establishing links between ontologies while the merging creates a new single ontology. The both approaches (i.e., horizontal and vertical) have different characters. The horizontal approach is used to analyze mapping between ontologies through integrated similar concept in the same level. The horizontal approach checks the relationships between the concepts of the same level in the two ontologies and merges or ties them as defined by WordNet. The vertical approach completes the merging operation for concepts with different levels, but belonging to the same branch of the tree. In this case they fill the resulting ontology with concepts from both ontologies. A similarity measure is calculated in order to define the hierarchy between these concepts in the resulting tree. While this method doesn't provide an adequate solution to automation, it provides a purely semantic approach to the merging solution.

#### 2) C.R. Rene Robin and G.V. Uma [8]

C.R. Rene Robin and G.V. Uma proposed an algorithm used for merging of ontologies automatically using a hybrid strategy. It consists of four sub strategies such as Lexical Matching, Semantic Matching, Similarity Check and Heuristics Functions. The user's only job is to give the OWL files as input and a merged file will be produced as output. Merging process starts from the top in one owl file and from bottom in other. They consider that ontologies completely differ if the leaf node has no similarity with the super most class. For comparing class names, Lexical Analysis and Semantic is used. Semantic Analysis uses WordNet as a database to identify the synonyms of the class names. If this has a match it means that the classes are same. So after a Similarity Check Heuristic is called. If Lexical and Semantic fails, it checks every class of the OWLfile1 with the class of OWLfile2 and saves the value of the classes and an intermediate value (the ratio of similarity between two classes). Similarity Checking of properties takes two classes as input, each of their properties is stored in an array. Every property of the class is compared with the other. To perform the comparison again lexical and semantic analysis is used. If the lexical and semantic match is found, the heuristic function is called. This process is repeated for every class of OWLfile2. In this process the output file OWLfile3 is initialized with owlfile1 any addition from OWLfile2 is made in OWLfile3. As in merging it should have all values that are in both OWL files. At last the owlfile3 is returned as merged file.

#### 3) N. Maiz and M. Fahad [9]

N. Maiz and M. Fahad presented a strategy for ontology merging in context of data warehousing by mediation that aims at building analysis contexts on-the-fly. Their methodology is based on the combination of the statistical aspect represented by the hierarchical clustering technique and the inference mechanism. Their approach on ontology merging topic has two folds. First, the semantic based ontology merger, Disjoint Knowledge Preservation based Ontology Merger (DKP-OM) system, follows the hybrid approach and uses various inconsistency detection algorithms in initial mapping found in first steps [10]. Their hybrid strategy makes it possible to find all possible mappings, and semantic validation of mappings gives very promising final results by ignoring the incorrect correspondences. In this approach the methodology starts by aligning the local ontologies to find similar entities belonging to different ones, the similarity between entities based on Wordnet thesaurus. Then, the result of the ontology alignment is used to merge local ontologies automatically. It generates the global ontology by four steps. First, it builds classes of equivalent entities of different categories (concepts, properties, instances) by applying a hierarchical clustering algorithm. Secondly, it makes inference on detected classes to discover new axioms representing the new relationships between entities in the same class or between different classes of the same category, and solves synonymy and homonymy conflicts. This step also consists of generating sets of concept pairs from ontology hierarchies, such as the first component subsumes the second one. Third, it merges different sets together, and uses classes of synonyms and sets of concept pairs to solve semantic conflicts in the global set of concept pairs. Finally, it transforms this set to a new hierarchy, which represents the global ontology. In this approach, it requires human intervention for the validation of mappings and semantic inconsistencies are neglected during the generation of merged ontology.

#### 4) M. Fahad and N. Molla [11]

M. Fahad and N. Moalla proposed an approach minimizes human involvement one step more down during the ontology merging process and it presented a novel methodology for the detection of semantic inconsistencies in the early stages of ontology merging. Disjoint knowledge analysis and preservation in ontology merging helps to identify the conceptualization mismatches between heterogeneous ontologies and provides more accuracy to the process of mapping. This results in global merged ontology free from 'circulatory error in class/property hierarchy', 'common class/instance between disjoint classes error', 'redundancy of subclass/ sub property relations', 'redundancy of disjoint relations' and other types of 'semantic inconsistency' errors. In this way, their methodology saves time and cost of traversing local ontologies for the validation of mappings, improves performance by producing only consistent accurate mappings, and reduces the user dependability for ensuring the satisfiability and consistency of merged ontology. This approach presented *Disjoint Knowledge Preservation based automatic Ontology Merger* (DKP-AOM) semantic-based ontology merger which extends the methodology of semi-automatic *DKP-OM* system [10] to encounter more structural and semantic conflicts, and to provide more optimized fully automatic solution for accessing and resolving semantic consistencies in an ontology merging. The newer developed DKP-AOM system employs various algorithms to detect inconsistent mappings from the initial list of mappings, and saves time and resources for the generation of hidden intermediate global ontology.

### 5) F. Freire de Araújo and F. Lígia Lopes [12]

F. Freire de Araújo and F. Lígia Lopes proposed an approach for automatic merging of multiple ontologies, called MeMO, which uses clustering techniques [13] in order to help the identification of the most similar ontologies. They consider that two important tasks must be executed to produce the global ontology: the similarity matrix building and the progressive ontology combination. MeMO approach produces a binary tree whose leaf nodes denote the source ontologies, and the root node represents the global ontology. The intermediary nodes represent the integrated ontologies, and are obtained during the merging process. One important aspect of MeMO is that it produces a global ontology which is really close to the ideal one.

### 6) Salvatore Raunich and E. Rahm [14]

Salvatore Raunich and E. Rahm demonstrated a new automatic approach to merge large taxonomies based on an equivalence matching between a source and target taxonomy to merge them. It is target-driven, i.e. it preserves the structure of the target taxonomy as much as possible. Furthermore, this approach can utilize additional relationships between source and target concepts to semantically improve the merging. They implemented Automatic Target-Driven Otology Merging (ATOM) system which is a new approach for taxonomy merging. It generates a default solution in a fully automatic way that may interactively be adapted by users if needed. ATOM base algorithm takes as input two taxonomies and an equivalence matching between concepts.

### B. Ontology Management Tools/Systems

This section provides an overview of the main ontology management tools that have been developed in the last years. These tools usually provide a graphical user interface for merging ontologies. They usually need the participation of the user to obtain the definitive result of the merging process. In the final of this section the comparison of these tools is included.

### 1) FCA-Merge [15]

FCA-Merge is a semi automatic tool for ontology merging based on Ganter and Wille's formal concept analysis [16], lattice exploration, and instances of ontologies to be merged. FCA Merge employs bottom up approach. The overall process of ontology merging consists of three steps:

a) Linguistic processing of source documents for extraction of instances and formal context.

b) Derivation of common context and computation of concept lattice.

c) The non automatic generation of the merged ontology with human interaction based on the concept lattice. Disadvantage of the technique is, it does not mention if the ontologies have the same/similar/complementary/orthogonal subject [8].

### 2) PROMPT [17]

PROMPT suite contains of a set of tools that had an important impact in the area of merging, aligning and versioning of ontologies. The suite includes an ontology merging tool (iPROMPT, formerly known as PROMPT), an ontology alignment tool (Anchor PROMPT), an ontology versioning tool (PROMPT Diff) and a tool for factoring out semantically complete sub-ontologies (PROMPTFactor). The PROMPT ontology merging algorithm begins with the linguistic-similarity matches for the initial comparison, generates a list of suggestions for the user based on linguistic and structural knowledge and then points the user to possible effects of these changes. This tool guides the user to remove inconsistencies in ontologies by determining the conflicts in the ontologies and suggests solutions.

### 3) Chimaera [18]

Chimaera is an interactive ontology merging and diagnosis tool. It is used to create, browse, edit, merge and diagnose ontologies. This application builds on a system called Ontolingua. It makes users affect merging process at any point during merge process. Chimaera allows the users to map ontologies by suggesting terms which are possible candidates in the ontologies to be merged or have taxonomic relationships which are to be included in the merged ontology. In fact, it is similar to PROMPT, as both are embedded in ontology editing environments and offer the user interactive suggestions. It solves mismatches at terminological and scope of concept level, and it helps alignment by providing possible edit points and it is not repeatability. But it is not automatic which means everything requires user interaction.

### 4) SAMBO (System for Aligning and Merging Biomedical Ontologies) [19]

SAMBO is a web-based ontology alignment and merging tool system, developed at Linköpings universitet in Sweden. SAMBO supports ontologies, which are represented in DAML+OIL and OWL and it is designed to allow two merging types:

- Suggestion Merge: Suggestions for possible merges are created by SAMBO by comparing the names and synonyms of slots and classes in the two ontologies but the user chooses which suggestions to merge or not.

- Manual Merge: The user can choose which slots and classes to merge without any suggestions from SAMBO and each merged item is added individually to the new ontology. The slot and class definitions that are not merged are copied to the new ontology. SAMBO provides a number of reasoning services such as consistency, satisfiability and equivalence checking by using the FaCT reasoner.

*5) HCONE [20]*

The goal is to validate the mapping between ontologies and to find a minimum set of axioms for the new merged ontology. Linguistic and structural knowledge about ontologies are exploited by the Latent Semantics Indexing method (LSI - a technique for information retrieval and indexing) [21] for associating concepts to their informal, human-oriented intended interpretations realized by WordNet senses. Using concept intended semantics; the proposed method translates formal concept definitions to a common vocabulary and exploits the translated definitions by means of description logics reasoning services. The HCONE approach is not completely automated; user involvements are placed at the early stages of the mapping/ merging process.

*6) iiMERGE (interactive merge) [22]*

iMERGE aimed to support the analytic comparing and merging process. It provides tightly linked and integrated techniques and views for visualizing. iMerge Provides a method for the user to accept/reject a suggested mapping, and allows access to full definitions of ontology terms, it Provides progress feedback on the overall mapping process. iMerge proposes some mappings but does not consider possible conflicts which may occur if the concepts are merged. Mapping and merging strategies in iMerge exploit the linguistic approach. With the method EditDistance [23] string similarity is computed from the number of edit operations (insertions, deletions and substitutions of single characters). It is necessary to transform one string into another one. Here, strings are compared according to their set of n-grams [24], i.e., sequences of n characters. The similarity between concepts based on their terminological relationships such as synonymy, hypernymy and hyponymy. This tool requires the use of auxiliary sources, such as documents or annotations.

*7) MOA [25]*

MOA is OWL ontology merging and alignment tool for semantic web based on a linguistic-information. Linguistic information is usually used to detect (dis)similarities between concepts. Many of them are based on syntactic and semantic heuristics such as concept name matching (e.g., exact, prefix, suffix matching). It works in most cases, and does not work well in more complex cases. In MoA, the correlations between the source ontologies are saved to a text file, and this can be viewed and edited in an editor. So, the users can accept or reject the result by editing the file. The intermediate output of MoA is a set of articulation rules between two ontologies; these rules define what the (dis)similarities are [26]. The final output of MoA (i.e., new mereged ontology) is similar to that of Chimaera and PROMPT. MoA refers not only to classes but also to slots and value restrictions. The merging algorithm of MoA accepts three inputs: two source ontologies and the semantic bridge ontology. MoA system is composed of four main components namely MoA engine, the core module of the architecture, is joined by Bossam inference engine, Ontomo editor, and Shell. They provide querying, editing, and user interfaces. Table 1 contains the information of the different tools used to management ontologies.

TABLE I.    THE COMPARISON BETWEEN MAIN FEATURES OF ONTOLOGY MANAGEMENT TOOLS

| Tool Name | Input | Output | Ontology Management Operation | Mapping strategy or algorithm | Automation Level | User interaction | auxiliary |
|---|---|---|---|---|---|---|---|
| FCA-Merge | Two ontologies and a set of documents of concepts in ontologies | Merged Ontology | Ontology merging | Linguistic analysis & TITANIC algorithm for computation for pruned concept lattice | Semi-automatic | Generating a merged ontology requires human interaction of the domain expert with background knowledge | Not Available |
| PROMPT | Two ontologies | Merged ontology | Merging and Alignment | Heuristic based analyzer | Semi-automatic | The user can accepts, Rejects or Adjusts system's suggestions. | Not Available |
| Chimara | RDF and DAML ontologies | Merged ontology | Ontology merging & diagnosing | Linguistic matcher | Semi-automatic | User can map ontologies by suggesting terms which are possible candidates in the ontologies to be merged | Not Available |
| SAMBO | Two ontologies | Merged ontology | Ontology merging and Alignment | Linguistic matcher | Semi-automatic | User can choose suggestions of merge system | Wordnet , UMLS |
| HCONE | Two ontologies | Merged ontology | Merging and Alignment | LSI | Semi-automatic | - | Wordnet |
| iMERGE | ontologies and a set of documents linked with the concepts | Merged ontology | Ontology merging | Linguistic approach with the method EditDistance and N-Gram | Semi-automatic | user can accept/reject a suggested mapping | Not Available |
| MOA | Two ontologies and semantic bridges | Merged ontology | Ontology merging and alignment | Linguistic analysis | Semi-automatic | Users can accept or reject the result by editing the output file | Wordnet |

The next section will discuss ontology merging methodology for building a single global ontology from heterogeneous XML data sources.

## III. ONTOLOGY BASED INTEGRATED FRAMEWORK

Because of the scattered nature of the Web, Information sources can be scattered in different XML data sources. Each information source should be mapped to its own local ontology. For this reason, the existence of an integrated framework for storing, querying and managing distributed XML data sources is of great importance. Towards an ontology based integrated framework, two modules are implemented, each of them aims to performing a specific task (Fig. 1):

1) **Automatic generation of local OWL ontologies module:** it is used to specify XML-to-OWL mappings for automatic generating local OWL ontologies from heterogeneous XML data sources. This module is implemented and described in [5]. It is composed of Jena, Trang, XSOM and JUNG as shown in Fig. 1.

2) **Merging Module:** it uses PROMPT framework for merging the local ontologies which are generated from previous module to build a global ontology covering the domain knowledge presented in XML data sources. This module will be described in details in the next section.

An ontology based integrated framework is written in Java and it uses several online-available APIs such as, Jena, Trang, XSOM, JUNG and PROMPT as shown in Fig. 1.

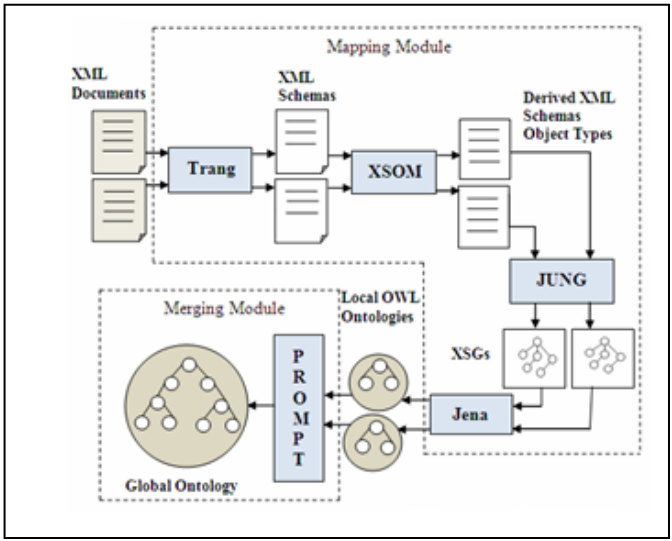

Figure 1. Ontology based integrated framework

## IV. ONTOLOGY MERGING

This section presents a process of merging different local ontologies for certain domain. These local ontologies are generated from heterogeneous XML data sources using system in [5]. The merging process is done through the protégé-owl and protégé frames where Prompt is plug-in/tab. The Prompt framework includes a tool called iPrompt which is responsible for merging ontologies.

### A. The IPROMPT Algorithm

The IPROMPT algorithm [17, 2] takes as input two ontologies and guides the user in the creation of one coherent merged ontology as output. Fig. 2 illustrates the iPrompt ontology-merging algorithm. First IPROMPT creates an initial list of matches based on lexical similarity of class names then the process goes through the following cycle (Fig. 2):

(1) The user triggers an operation by either selecting one of IPROMPT's suggestions from the list or by using an ontology-editing environment to specify the desired operation directly.

(2) IPROMPT performs the operation, automatically executes additional changes based on the type of the operation, generates a list of suggestions for the user based on the structure of the ontology around the arguments to the last operation, and determines inconsistencies and potential problems that the last operation introduced in the ontology and finds possible solutions for those problems.

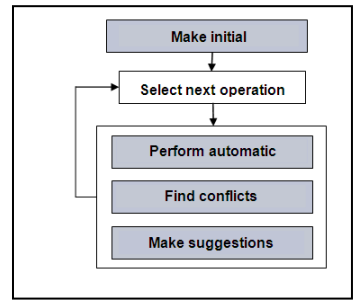

Figure 2. The workflow of IPROMPT algorithm. The colored boxes indicate the actions performed by IPROMPT. The white box indicates the action performed by the user.

### B. The IPROMPT Ontology Merging Operations

The set of ontology-merging operations includes both the operations that are normally performed during traditional ontology editing of a single ontology and the operations specific to merging and alignment, such as merging classes, merging slots, merging instances, performing a deep or a shallow copy of a class from one ontology to another [17]. In the descriptions of the operations below, $O_m$ is the merged Ontology.

**Merge classes:** to merge two classes A and B, create a new class M in $O_m$. For each class C that is a subclass or a super class of A or B, if there is an image $C_i$ of C in $O_m$, $C_i$ becomes a subclass or a super class of M, respectively. For each slot S attached to A or B, if there is no image of S in $O_m$, copy S to $O_m$. For each image $S_i$ of S, attach $S_i$ to the class M. If either A or B was already in Om prior to the operation, all references to it in $O_m$ become references to M and the original frame is deleted.

**Merge slots:** to merge two slots $S_1$ and $S_2$ create a new slot $S_m$. For each class C in the domain and range of $S_1$ or $S_2$, if there is an image $C_i$ of C in $O_m$, add $C_i$ to the domain or range of $S_m$, respectively. If either $S_1$ or $S_2$ was in $O_m$ prior to the operation, all references to it become references to $S_m$ and the original slot is deleted. Note that as a result of the last step, if there was a class in $O_m$ that had both $S_1$ and $S_2$ attached to it, after the operation it will have only $S_m$ attached to it. In addition, IPROMPT suggests that the user merges classes in the domain (and range) of $S_m$ that came from different source ontologies. For instance, if there exist in one of the source ontologies, a slot sex had the class Gender as its range and in the other a slot sex had a class Sex as its range. If the user merges the two sex slots, IPROMPT will suggest merging classes Gender and Sex.

**Merge instances:** IPROMPT does the following when merging two instances $I_1$ and $I_2$ to create a new instance $I_m$: If classes $C_1$ and $C_2$ which are the types of $I_1$ and $I_2$ respectively have no images in $O_m$, copy them to $O_m$ (see the operation perform a shallow copy of a class). If $C_1$ and $C_2$ already have images in $O_m$ and the images are different frames, merge the images (the user must confirm this operation). Note that as a result, the merged instance $I_m$ will have the same slots, or their images, that $I_1$ and $I_2$ had. For each value V for each slot S attached to $I_1$ or $I_2$, do the following:

- If V is a primitive value (string, number, etc.), add V to the value of the image of S for $I_m$.
- If V is a frame and there is an image of V, $V_i$, in Om, add $V_i$ to the value of the image of S for $I_m$. For each slot S of $I_m$ that has values that are images of frames coming from different sources, IPROMPT suggests merging these frames. Note that adding images of all the values at sources can create violations of range and cardinality constraints for the merged instance. This inconsistency is one of the inconsistencies that IPROMPT checks for in the next step of the algorithm - finding inconsistencies and potential problems.

**Perform a shallow copy of a class:** copy a class from a source ontology to another. When copying a class C, create a new class $C_i$ in $O_m$. For each slot S directly attached to C, if there is no image of S in $O_m$, copy S to $O_m$. Then attach images of all the slots of C to $C_i$.

**Perform a deep copy of a class:** copy a class from one ontology to another copying all the parents of a class up to the root of the hierarchy. To perform a deep copy of a class C, perform its shallow copy, and then perform a deep copy for each of its super classes.

In terms of user support, the IPROMPT tool has the following features [17]:

1- Setting the preferred ontology: It often happens, that the source ontologies are not equally important or stable, and that the user would like to resolve all the conflicts in favor of one of the source ontologies.

IPROMPT allows the user to designate one of the ontologies as preferred. When there is a conflict between values, instead of presenting the conflict to the user for resolution, the system resolves the conflict automatically.

2- Maintaining the user's focus: Suppose a user is merging two large ontologies and is currently working in one content area of the ontology. IPROMPT maintains the user's focus by rearranging its lists of suggestions and conflicts and presenting first the items that include frames related to the arguments of the latest operations.

3- Providing feedback to the user: For each of its suggestions, IPROMPT presents a series of explanations, starting with why it suggested the operation in the first place. If IPROMPT later changes the operation placement in the suggestions list, it augments the explanation with the information on why it moved the operation.

The next section describes experimental results of proposed ontology-based integrated framework after applying the merging process on the generated local ontologies which are described in [5].

## V. EXPERIMENTAL RESULTS

This section includes the description of the experiments that resulted from ontology merging environments.

Fig. 3 shows the two local ontologies used for the experiment. The two local ontologies have been generated from heterogeneous XML data sources using the automatic generating of local OWL ontologies module which is described in [5]. Both ontologies represent structure of scientific publications. They were encoded in OWL-Description Logic (DL) format [27]. Fig. 3-a shows "Ruby_bibliography" Ontology which composes of four OWL classes, named: bibliography, biblioentry, author and publisher. While Fig. 3-b shows "Niagara_bib" Ontology which composes of four OWL classes named: bib, vendor, book and author. Various screenshots are showed to discuss how two ontologies are merged using PROMPT tab in protégé.

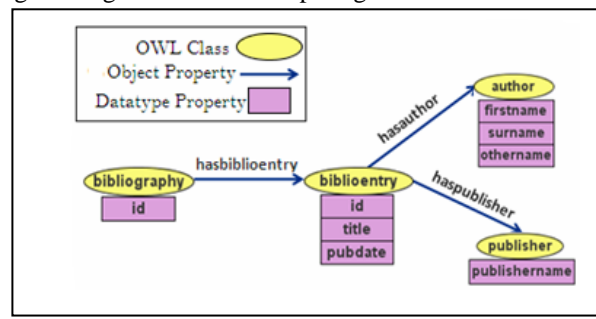

*(a)*

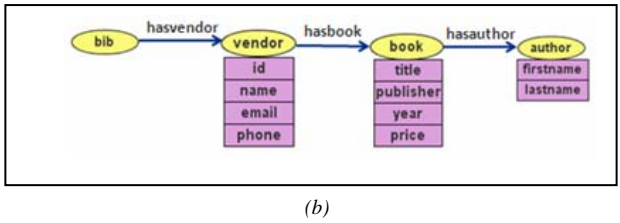

*(b)*

Figure 3.    The two local ontologies. (a) Ruby_bibliography Ontology; (b) Niagara_bib Ontology References.

In order to proceed for merging the two local ontologies presented above, PROMPT tab in protégé is used. After loading the sources ontologies (local ontologies) and performing an initial analysis of class names in IPROMPT. It then displays the results of its analysis at the Suggestions tab. The suggestions screen is obtained as shown in Fig. 4.

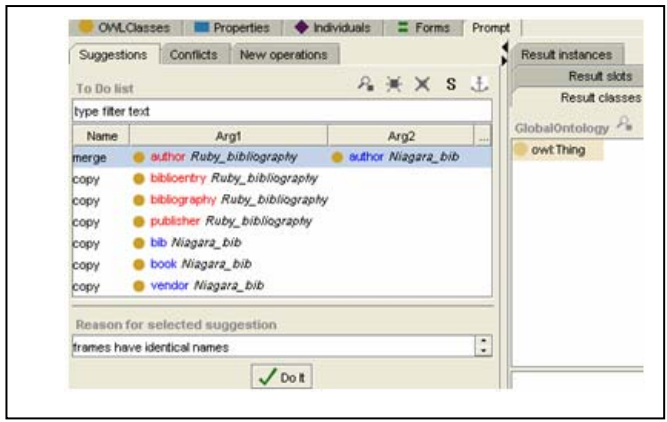

Figure 4.    Suggestion shown by PROMPT tab for merging two local ontologies.

As discussed in section 3.2, in order to merge two classes. For example, the user can follow the IPROMPT's suggestion list and merge two author classes. Also, the user can define new operation merge-classes to merge "bibliography" and "bib" classes in the Sources window as shown in Fig. 5. IPROMPT creates automatically new class named "bibliography" in the GlobalOntology.

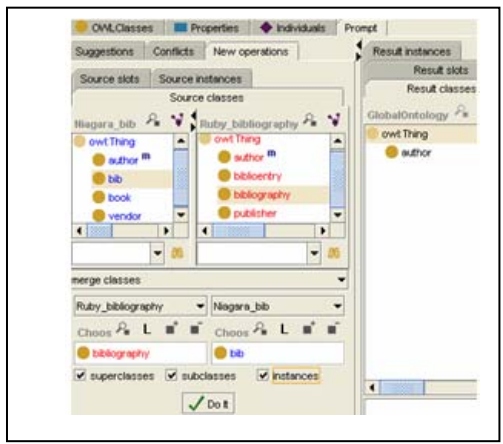

Figure 5.    Source classes of two ontologies being merged in defined new operations.

After copying the remained classes of two sources ontologies from the IPROMPT suggestion list such as biblioentry, publisher, book and vendor, the resulting classes of the merged ontology (GlobalOntology) are obtained as shown in Fig. 6. Note that some of the classes in the merged ontology have the suffix "Ruby_bibliography", some of them have the suffix "Niagara_bib" and the rest have no suffix. The classes that have no suffix describe the shared classes between the two ontologies which are created by merging the similar classes in the two ontologies. While the classes having suffix "Ruby_bibliography" represent the classes that are copied from the "Ruby_bibliography" ontology and doesn't exist in the other ontology and in the same way the classes that have the suffix "Niagara_bib" represent the classes that are copied from the ontology "Niagara_bib and doesn't exist in the other ontology.

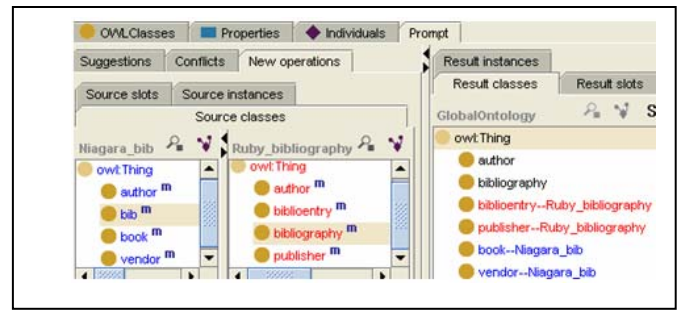

Figure 6.    The result classes of the merged ontology.

As discussed in section 3.2, in order to merge two slots. In our experiment, the system suggests merging the two slots firstname. We define new operations to merge two slots such as othername and lastname to obtain new merged slot named lastname. Then copying the remained slots like id, pubdate, name, publishername, publisher, email, phone, title, year, price, has biblioentry, hasbook and hasvendor.

Protégé allows us to modify the ontology structure. So, we can rename or remove ontology terms, or change the domain and the range of a property, or create new classes. In our case, the two classes "bibliography" and "author" are extended, with Publication and Person being their superclasses, respectively. Fig. 7 shows the new merged ontology after merging the two ontologies and restructuring the new merged ontology.

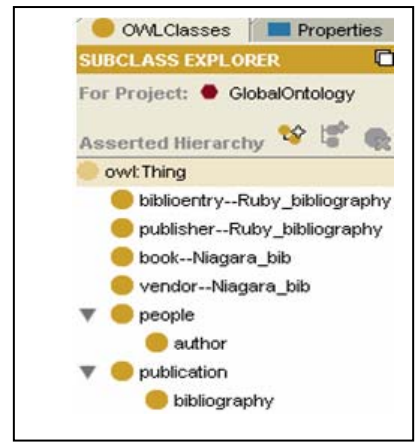

Figure 7. Classes hierarchy of the new merged ontology.

Tables 2 and 3 show object properties and datatypes properties of the new merged ontology in order.

TABLE II.    THE OBJECT PROPERTIES OF GLOBAL ONTOLOGY

| Object Properties | Domain | Range |
|---|---|---|
| hasbiblioentry | bibliography | biblioentry |
| hasvendor | bibliography | vendor |
| hasbook | vendor | book |
| hasauthor | book | author |
| hasauthor | biblioentry | author |
| haspublisher | biblioentry | publisher |
| hasissue | SigmodRecord | issue |
| hasarticles | Issue | articles |
| hasarticle | Articles | article |
| hasauthors | Article | authors |
| hasauthor | Authors | author |

TABLE III.    HE DATATYPE PROPERTIES OF GLOBAL ONTOLOGY

| Datatype Properties | Domain | Range |
|---|---|---|
| id | bibliography | xsd:NCName |
|  | biblioentry | xsd:NCName |
|  | vendor | xsd:NCName |
| name | vendor | xsd:NCName |
| email | vendor | xsd:string |
| phone | vendor | xsd:NMTOKEN |
| title | book | xsd:string |
| publisher | book | xsd:string |
| year | book | xsd:integer |
| price | book | xsd:decimal |
| title | biblioentry | xsd:string |
| pubdate | biblioentry | xsd:integer |
| volume | issue | xsd:integer |
| number | issue | xsd:integer |
| title | article | xsd:string |
| initPage | article | xsd:integer |
| endPage | article | xsd:integer |
| position | author | xsd:integer |
| firstname | Author | xsd:NCName |
| surname | author | xsd:NCName |
| lastname | author | xsd:NCName |

## VI.    CONCLUSIONS AND FUTURE WORK

In this paper, some issues of ontology management process like methodologies and tools have been highlighted. It illustrates with an example, the merging process, which is the second module of ontology based integrated framework. IPROMPT algorithm identifies potential merge candidates based on class-name similarities. The result is presented to the user as a list of potential merge operations then the user chooses one of the suggested operations from the list or specifies the operation directly. The system performs the requested action and automatically executes additional changes derived from the action. It then makes a new list of suggested actions for the user based on the new structure of the ontology, determines conflicts introduced by the last action, finds possible solutions to these conflicts and displays these to the user. From this example, it is clear that the merging process requires the intervention of human to be done without conflicts which means that the fully automation of merging process is almost impossible since it requires good knowledge of the domain, understanding of each ontology point of view, and even the use of negotiation strategies between the designers of the different ontologies in order to make proposals, discuss them and to reach an agreement. In this paper, data instances are not materialized at the local ontologies. The global ontology only contains the concepts and properties but not the instances, which stay in the source and are retrieved and translated as needed in response to user queries. For this reason, the subsequent work will be focused on query translation process within this system for XML data sources.